\documentclass[a4paper,conference]{IEEEtran}

\usepackage[dvipsnames, table]{xcolor}         
\usepackage{hyperref}
\usepackage{url}
\usepackage[pdftex]{graphicx}

\makeatletter
\let\MYcaption\@makecaption
\makeatother

\usepackage[font=footnotesize]{subcaption}

\makeatletter
\let\@makecaption\MYcaption
\makeatother

\usepackage[percent]{overpic}
\usepackage{physics}
\usepackage{booktabs}  
\usepackage{amsfonts}  

\definecolor{good}{HTML}{cccccc}

\graphicspath{{images/}}

\begin{document}
%
\title{Autoencoder for Synthetic to Real Generalization: From Simple to More Complex Scenes}

\author{\IEEEauthorblockN{Steve Dias Da Cruz\IEEEauthorrefmark{1}\IEEEauthorrefmark{2}\IEEEauthorrefmark{3},
Bertram Taetz\IEEEauthorrefmark{3},
Thomas Stifter\IEEEauthorrefmark{1},
Didier Stricker\IEEEauthorrefmark{2}\IEEEauthorrefmark{3}}
\IEEEauthorblockA{\IEEEauthorrefmark{1}IEE S.A.}
\IEEEauthorblockA{\IEEEauthorrefmark{2}University of Kaiserslautern}
\IEEEauthorblockA{\IEEEauthorrefmark{3}German Research Center for Artificial Intelligence\\
}
\IEEEauthorblockA{Email: steve.dias-da-cruz@iee.lu, bertram.taetz@dfki.de, thomas.stifter@iee.lu, didier.stricker@dfki.de}}

\maketitle

\begin{abstract}
  Learning on synthetic data and transferring the resulting properties to their real counterparts is an important challenge for reducing costs and increasing safety in machine learning. In this work, we focus on autoencoder architectures and aim at learning latent space representations that are invariant to inductive biases caused by the domain shift between simulated and real images showing the same scenario. We train on synthetic images only, present approaches to increase generalizability and improve the preservation of the semantics to real datasets of increasing visual complexity. We show that pre-trained feature extractors (e.g. VGG) can be sufficient for generalization on images of lower complexity, but additional improvements are required for visually more complex scenes. To this end, we demonstrate a new sampling technique, which matches semantically important parts of the image, while randomizing the other parts, leads to salient feature extraction and a neglection of unimportant parts. 
  This helps the generalization to real data and we further show that our approach outperforms fine-tuned classification models. 
\end{abstract}


\IEEEpeerreviewmaketitle

\section{Introduction}
The generation of synthetic data constitutes a cost efficient way for acquiring machine learning training data together with exact and free annotations. Notwithstanding this obvious advantage, bridging the gap between synthetic and real data remains an open challenge, in particular for camera based applications. Learning from synthetic data is an important tool in robotics: for example, to train a quadrupedal robot on synthetic data by incorporating proprioceptive feedback \cite{Leeeabc5986}, to train a robot hand to solve real Rubik's cubes by learning the model in a simulation only \cite{akkaya2019solving} or by translating the real world input data into synthetic data for a reinforcement learning agent \cite{zhang2019vr} and to \textit{make  the  robot  feel  at  home}. In view of safety critical applications, synthetic data can provide the means to reduce costs related to acquiring samples for edge cases, or which are difficult to obtain since they are too dangerous, e.g. accidents. We focus on learning invariances empirically on synthetic data, which should transfer to real data, opposed to constructing invariances as in equivariant neural networks \cite{Romero2020CoAttentive}.

We investigate the case of single independent images for which consistency between frames and physical interactions cannot be taken advantage of. The latter is commonly used by reinforcement learning methods \cite{Leeeabc5986}. 
We focus on training on synthetic data only and limit ourselves to autoencoder models which provide interesting properties due to their bottleneck design. 
The low-dimensional latent space of autoencoders can be subject to metric constraints \cite{hoffer2015deep}, allows for scene decomposition \cite{Engelcke2020GENESIS} and it is believed that latent factor disentanglement can be useful for downstream tasks \cite{NEURIPS2019_bc3c4a63}.
We assess to what extend we can generalize to real images and we highlight which design choices improve the autoencoder models performance with respect to accuracy and reconstruction quality.
To this end, we first develop a method using features of pre-trained classifiers and show that we achieve better results on MPI3D \cite{MPI3D} to generalize from synthetic (toy or realistic) to real images compared to Autoencoder, Variational Autoencoder (VAE) \cite{Kingma2014Autoencoding}, $\beta$-VAE \cite{Higgins2017Beta} and FactorVAE \cite{kim2018disentangling}. 
Although successful, we highlight that insights and design choices on a simple dataset do not necessarily transfer to real applications of higher visual complexity. 
To improve generalization, we propose to use the partially impossible reconstruction loss (PIRL) \cite{DiasDaDa2021Illumination} (matching semantically important parts while randomizing the other parts) and we propose a novel variation thereof.  We extensively show that our variation is the driving force for the improved generalization capacities. Additionally, we induce structure in the latent space by a triplet loss regularization. 
We evaluate and justify the benefits of the different design choices on an automotive application focusing on occupancy classification in the vehicle interior. 
The challenge of training in a single vehicle interior and transferring results between different vehicle interiors has been investigated \cite{dias2021iv}. The latter and similar industrial applications suffer from the limited availability and variability of training data.
A successful transfer from synthetic to real data would avoid the necessity of collecting real data for each vehicle interior: the invariances could be learned and improved on synthetic data only.
\begin{figure*}
  \centering
  \includegraphics[width=0.80\linewidth]{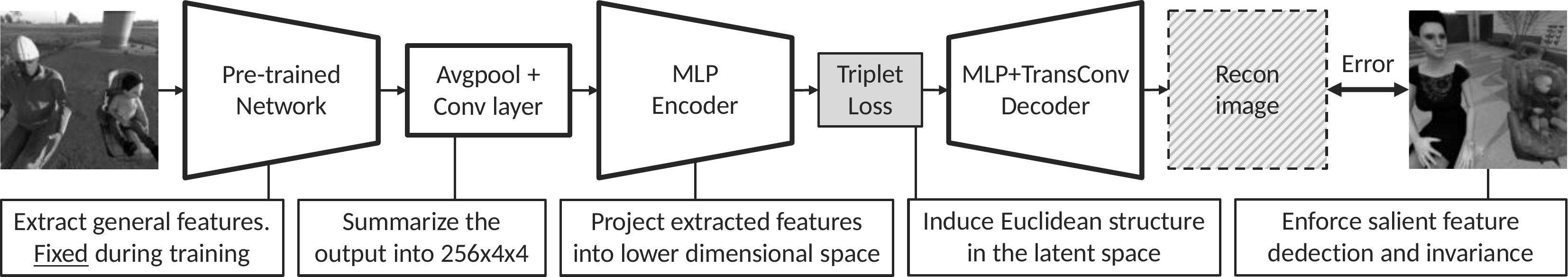}
  \caption{Impossible Instance Extractor Triplet Autoencoder (II-E-TAE) model architecture.}
  \label{fig:architecture}
\end{figure*}

\section{Related Works}
There have been successful applications of reinforcement learning systems being trained in a simulated environment and deployed to a real one, for example by combining real and synthetic data during training \cite{kang2019generalization, Rao_2020_CVPR, fang2018multi, bewley2019learning}. However, these approaches can take into account temporal information and action-reaction causalities while in this work we use independent frames only. A good overview on reinforcement learning based simulation to real transferability is provided in \cite{zhao2020sim}. Another line of research uses generative adversarial networks (GAN) to make synthetic images look like real images or vice versa \cite{ho2020retinagan, carlson2019sensor}. This requires both synthetic and real images, whereas we focus on training on synthetic images only. Part of our methodology is related to domain randomization \cite{tremblay2018training}, where the environment is being randomized, but the authors deployed this to object detection and the resulting model needs to be fine-tuned on real data. A similar idea of freezing the layers of a pre-trained model was investigated for object detection \cite{hinterstoisser2018pre}, but neither with a dedicated sampling strategy nor in the context of autoencoders. Another work focuses on localization and training on synthetic images only \cite{tobin2017domain}, though the applicability is only tested on simple geometries. Although, we start our investigations on the simple dataset MPI3D, we increase the visual complexity by incorporating human models and child seats. Others rely on the use of real images during training for the minimization of the synthetic to real gap for autoencoders \cite{inoue2018transfer, zhang2015learning}. Recent advances on synthetic to real image segmentation \cite{chen2020automated, yue2019domain, pan2018two} on the VisDA \cite{visda2017} dataset show a promising direction to overcome the gap between synthetic and real images, however, this cannot straightforwardly be compared against the investigation in this work, particularly, since we are focusing on autoencoder models and their generative nature. While our cost function variation is based on a previous work \cite{DiasDaDa2021Illumination}, we show that our approach improves generalization while needing less demanding training data such that it can easily be applied to any commonly recorded classification dataset (i.e. no variations of the same scene are needed).

\section{Method}
\label{sec:method}
Consider $N_s$ sceneries and $N_v$ variations of the same scenery, e.g. same scenery under different illuminations, with different backgrounds or under different data augmentation transformations. Let $\mathcal{X}=\{X_i^j\, \vert\, 1 \leq i \leq N_v, 1 \leq j \leq N_s\}$ denote the training data, where each $X_i^j \in \mathbb{R}^{C \times H \times W}$ is the $i$th variation of scene $j$ consisting of $C$ channels and being of height $H$ and width $W$. Let $X^j=\{X_i^j\, \vert\, 1 \leq i \leq N_v\}$ be the set of all variations $i$ of scenery $j$ and $\mathcal{Y}=\{Y^j\, \vert\, 1 \leq j \leq N_s\}$ be the corresponding target classes of the scenes of $\mathcal{X}$. Notice that the classes remain constant for the variations $i$ of each scene $j$. In the following, we will present the final model architecture as illustrated in Fig. \ref{fig:architecture} and we provide evidences for each design choice in Section \ref{sec:experiments}.

\subsection{Model Architecture: Extractor Autoencoder}
By an abuse of terminology, we will refer to our method as a variation of vanilla autoencoders, although an encoder-decoder formulation would strictly speaking be more correct, because the goal will not be to reconstruct the input image exactly. We propose to apply ideas from transfer learning and use a pre-trained classification model to extract more general features from the input images. Instead of using the images itself, the extracted features are used as input. Our autoencoder consists of a summarization module which reduces the number of convolutional filters. This is fed to a simple MLP encoder which is then decoded by a transposed convolutional network. We refer to this model as \textit{extractor autoencoder} (E-AE). Let $\mathrm{e}_{\phi}$ be the encoder, $\mathrm{d}_{\theta}$ the decoder and $\mathrm{ext}_{\omega}$ be a pre-trained classification model, referred to as \textit{extractor}. For ease of notation, we define $\mathrm{e}_{\phi}(\mathrm{ext}_{\omega}(\cdot))=\mathrm{ee}_{\phi,\omega}(\cdot)$. The model, using the vanilla reconstruction loss, can be formulated as
\begin{align}
  \mathcal{L}_R(X_i^j; \theta, \phi) &= \mathrm{r}\left(\mathrm{d}_{\theta}(\mathrm{e}_{\phi}(\mathrm{ext}_{\omega}(X_i^j))), X_i^j\right) \\
  &= \mathrm{r}\left(\mathrm{d}_{\theta}(\mathrm{ee}_{\phi,\omega}(X_i^j)), X_i^j\right),
\label{eq:recon-loss}
\end{align}
where $\mathrm{r(\cdot, \cdot)}$ computes the error loss between target and reconstruction. We use the structural similarity index measure (SSIM) \cite{bergmann2018improving} and binary cross entropy (BCE). Model details are provided in the appendix Section S2-A.

\subsection{Sampling Strategy: Partial Impossible}
An additional improvement to the autoencoder training approach is a dedicated sampling strategy for which we provide two variations. The first one is the partially impossible reconstruction loss (PIRL) as introduced for illumination normalization \cite{DiasDaDa2021Illumination}. As our results will show, this also helps the transfer between synthetic and real images. For sampling the individual elements of a batch, we randomly select for each scene two images, one as input and the other one as target. This sampling strategy preserves the semantics while varying the unimportant features such that the model needs to focus on what remains constant. For random $a,b \in [0, N_v]$ and $a \neq b$:
\begin{equation}
  \mathcal{L}_{R,I}(X_a^j; \theta, \phi) = \mathrm{r}\left(\mathrm{d}_{\theta}(\mathrm{ee}_{\phi,\omega}(X_a^j)), X_b^j\right).
\label{eq:recon-loss-I}
\end{equation}
We refer to using the PIRL by prepending an \textit{I}, e.g. I-E-AE. 

\subsection{Sampling Strategy: Partial Impossible Class Instance}
\begin{figure}
  \centering
  \begin{subfigure}{0.15\textwidth}
      \centering
      \begin{overpic}[width=0.48\linewidth]{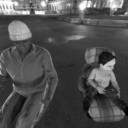}
        \put(30,80){\footnotesize{\textcolor{white}{Input}}}
      \end{overpic}
      \begin{overpic}[width=0.48\linewidth]{input_target_example/input_1.png}
        \put(25,80){\footnotesize{\textcolor{white}{Target}}}
      \end{overpic}
      \caption{Vanilla: Eq. \eqref{eq:recon-loss}}
      \label{fig:vanilla-recon}
  \end{subfigure}
  \hfill
  \begin{subfigure}{0.15\textwidth}
      \centering
      \begin{overpic}[width=0.48\linewidth]{input_target_example/input_1.png}
        \put(30,80){\footnotesize{\textcolor{white}{Input}}}
      \end{overpic}
      \begin{overpic}[width=0.48\linewidth]{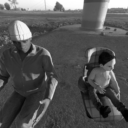}
        \put(25,80){\footnotesize{\textcolor{white}{Target}}}
      \end{overpic}
      \caption{I-PIRL: Eq. \eqref{eq:recon-loss-I}}
      \label{fig:impossible-recon}
  \end{subfigure}
  \hfill
  \begin{subfigure}{0.15\textwidth}
      \centering
      \begin{overpic}[width=0.48\linewidth]{input_target_example/input_1.png}
        \put(30,80){\footnotesize{\textcolor{white}{Input}}}
      \end{overpic}
      \begin{overpic}[width=0.48\linewidth]{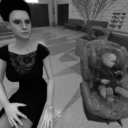}
        \put(25,80){\footnotesize{\textcolor{white}{Target}}}
      \end{overpic}
      \caption{II-PIRL: Eq. \eqref{eq:recon-loss-II}}
      \label{fig:instance-impossible-recon}
  \end{subfigure}
  \caption{Different input-target pairs for the reconstruction loss.}
  \label{fig:input}
\end{figure}
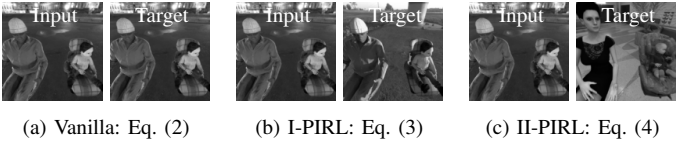

We propose a novel variation to further improve this strategy by sampling a target image of a different scene, but of the same class. This should cause the model to learn invariances with respect to certain class variations which are not important for the task at hand, e.g. clothes, human poses, textures. This sampling variation is reflected in the reconstruction loss by
\begin{equation}
  \mathcal{L}_{R,II}(X_a^j; \theta, \phi) = \mathrm{r}\left(\mathrm{d}_{\theta}(\mathrm{ee}_{\phi,\omega}(X_a^j)), X_b^k\right),
\label{eq:recon-loss-II}
\end{equation}
for random $a,b \in [0, N_v]$, $j \neq k$ and $Y^j=Y^k$. We refer to this method as impossible class instance sampling marked by prepending \textit{II}, e.g. II-E-AE. It is important to notice that our novel variation can easily be applied to any common dataset. The sampling variations are visualized in Fig. \ref{fig:input}.

\subsection{Structure in the Latent Space: Triplet Loss}
The final adjustment to our training strategy is the incorporation of the triplet loss regularization in the latent space \cite{hoffer2015deep} to induce structure. This can be integrated by 
\begin{align}
    \mathcal{L}_T(X_a^j;\phi) = &\max \left(0, \norm{\mathrm{ee}_{\phi,\omega}(X_a^j) -\mathrm{ee}_{\phi,\omega}(X_b^k)}^2 \right. \nonumber \\
    & \left. - \norm{\mathrm{ee}_{\phi,\omega}(X_a^j)-\mathrm{ee}_{\phi,\omega}(X_c^l)}^2 + 0.2 \right),
\label{eq:triplet-loss}
\end{align}
for random $a,b,c \in [0, N_v]$, $j \neq k \neq l$ and $Y^j = Y^k \neq Y^l$. We refer to this model as \textit{triplet autoencoder} (TAE) either with or without using the PIRL. We can sample impossible target instances for the positive and negative triplet samples such that the total loss becomes (for some  $\alpha$ and $\beta$):
\begin{align}
  \mathcal{L}(X_a^j;\theta, \phi) =  &\alpha \mathcal{L}_T(X_a^j;\phi) + \beta \left(\mathcal{L}_{R,II}(X_a^j;\theta, \phi) \right. \nonumber \\
  & \left. + \mathcal{L}_{R,II}(X_b^k;\theta, \phi) + \mathcal{L}_{R,II}(X_c^l;\theta, \phi)\right).
\label{eq:loss}
\end{align}

\section{Experiments}
\label{sec:experiments}
This section is organized in observations, formulated as subsections, which are built on one another and contain results highlighting the improvements. This provides explanations for the design choices leading to our final model architecture and cost function formulations presented in Section \ref{sec:method}. Improvements regarding the transfer to real images when only being trained on synthetic images are assessed qualitatively based on reconstruction quality and latent space structure and quantitatively on classification accuracy. All experiments use the same hyperparameters whenever possible. Training details are provided in the appendix and in \href{https://github.com/SteveCruz/icpr2022-autoencoder-syn2real}{our implementation (link)}.

We perform a baseline evaluation on \href{https://github.com/rr-learning/disentanglement_dataset}{MPI3D} \cite{MPI3D}, which provides simple and realistic renderings and real counterparts. We reduced the dataset to contain only the large objects. For a higher visual complexity, we use as synthetic images the \href{https://sviro.kl.dfki.de/}{SVIRO} \cite{DiasDaCruz2020SVIRO} dataset. \href{https://vizta-tof.kl.dfki.de/}{TICaM} \cite{katrolia2021ticam} is used to evaluate the performance on a real dataset of a similar application. The latter datasets are grayscale images from the vehicle interior and consider the task of classification (empty, infant, child or adult) for each seat position. The design choices made on MPI3D and the available synthetic images are not sufficient to obtain a good transferability to real images from the vehicle interior. Hence, we release an additional dataset, see Section \ref{sec:impossible} and S1-D in the appendix. We introduce step by step modifications to the autoencoder architecture leading to steady quantitative and qualitative improvements. MPI3D and the vehicle interior share interesting properties: they have almost identical backgrounds and the environment is more tractable than many computer vision datasets. The transfer from SVIRO to TICaM is further complicated by new unseen attributes, e.g. steering wheel. An additional ablation study shows that our novel variation of PIRL is the driving force for the improved generalization capacity. Finally, to be in line with common benchmark datasets, we show that our design choices also improve the transfer from training on MNIST \cite{lecun1998gradient} to generalizing to real images of digits \cite{de2009character}.  

\subsection{Autoencoders struggle on real images when trained on synthetic images}
\label{sec:vanilla}
In the first, albeit naïve experiment we assumed that due to the bottleneck of autoencoders, the latter should generalize to some extent to real images when trained on synthetic ones. We trained convolutional autoencoders (AE) on the toy and realistic MPI3D images, respectively, and evaluated the resulting models on the real recordings. The first row of Fig. \ref{fig:mpi3d-realistic} shows the reconstruction of real images when trained on the realistic synthetic images: the model preserves some of the semantics. The model fails to perform senseful reconstructions when trained on toy images, see Fig. \ref{fig:mpi3d-toy}.

\subsection{Autoencoders overfit to the synthetic distribution}
\label{sec:overfit}
A consequence of the results of the previous section is the assumption that the autoencoder overfits to the synthetic distribution and takes into consideration some artefacts (e.g. rendering noise). We followed the idea of the MPI3D authors \cite{MPI3D} and trained Variational Autoencoder (VAE) \cite{Kingma2014Autoencoding}, $\beta$-VAE \cite{Higgins2017Beta} and FactorVAE \cite{kim2018disentangling} on the same data as before using the BCE reconstruction loss. The results in the second ($\beta$-VAE with $\beta=8$) and third (FactorVAE with $\gamma=50$) row of Fig. \ref{fig:mpi3d-realistic} show that the models reconstruct real images better and more of the semantics are preserved. If trained on toy renderings, the representation gap is too large, causing the reconstruction of the real images to be bad: see Fig. \ref{fig:mpi3d-toy}. 

\subsection{More general input features improve reconstructions}
\label{sec:extractor}
A small gap between the synthetic and real distribution can potentially be closed by a dedicated data augmentation approach to avoid overfitting to synthetic artefacts. Nevertheless, an abstraction from toy to real images cannot be achieved by means of simple data transformations or model constraints (e.g. denoising autoencoder). To this end we propose to use a pre-trained feature extractor as presented in Section \ref{sec:method} and as defined by Eq. \eqref{eq:recon-loss}. We used the VGG-11 model pre-trained on Imagenet as the extractor if not stated otherwise.
\begin{figure}
\begin{subfigure}{0.50\textwidth}
    \centering
    \begin{overpic}[width=\linewidth]{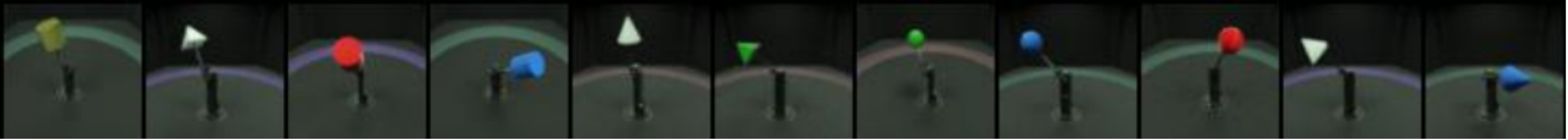}
      \put(0.6,1.2){\rotatebox{90}{\tiny{\textcolor{white}{Realistic}}}}
    \end{overpic}%
    \vspace{-0.07\baselineskip}
    \begin{overpic}[width=\linewidth]{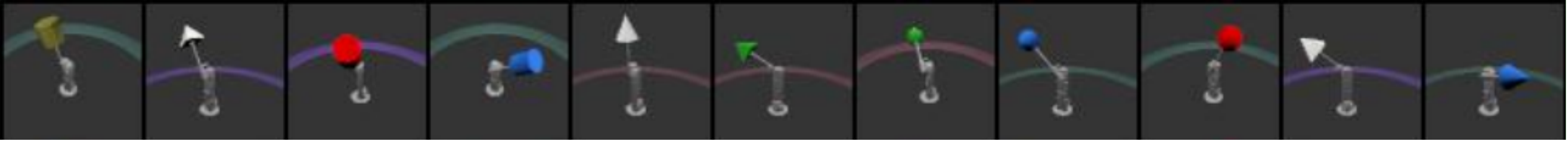}
      \put(0.6,3.0){\rotatebox{90}{\tiny{\textcolor{white}{Toy}}}}
    \end{overpic}%
    \vspace{-0.07\baselineskip}
    \begin{overpic}[width=\linewidth]{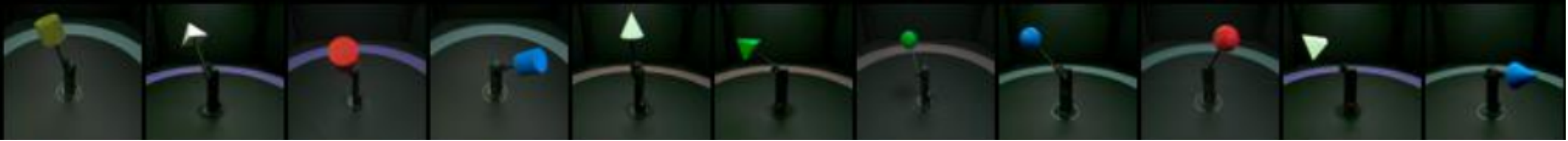}
      \put(0.6,2.5){\rotatebox{90}{\tiny{\textcolor{white}{Real}}}}
    \end{overpic}%
    \caption{Synthetic realistic and toy data used for training respectively, as well as real data used as input after training for evaluation.}
    \label{fig:mpi3d-data}
\end{subfigure}%
\vspace{0.5\baselineskip}
\begin{subfigure}{0.50\textwidth}
    \centering
    \begin{overpic}[width=\linewidth]{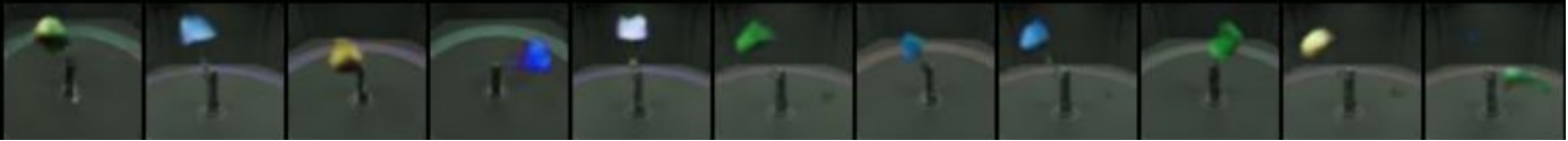}
      \put(0.6,2.9){\rotatebox{90}{\tiny{\textcolor{white}{AE}}}}
    \end{overpic}%
    \vspace{-0.07\baselineskip}
    \begin{overpic}[width=\linewidth]{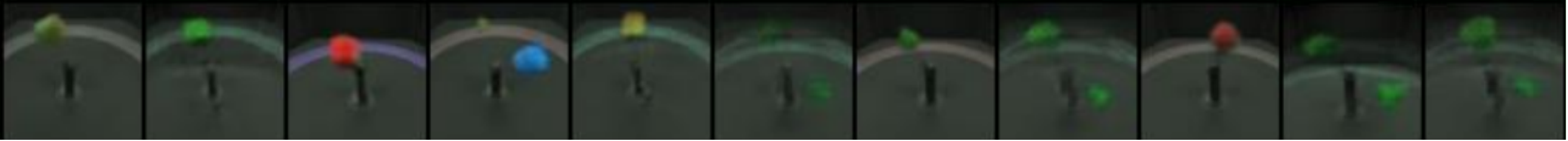}
      \put(0.6,1.9){\rotatebox{90}{\tiny{\textcolor{white}{$\beta$-AE}}}}
    \end{overpic}%
    \vspace{-0.07\baselineskip}
    \begin{overpic}[width=\linewidth]{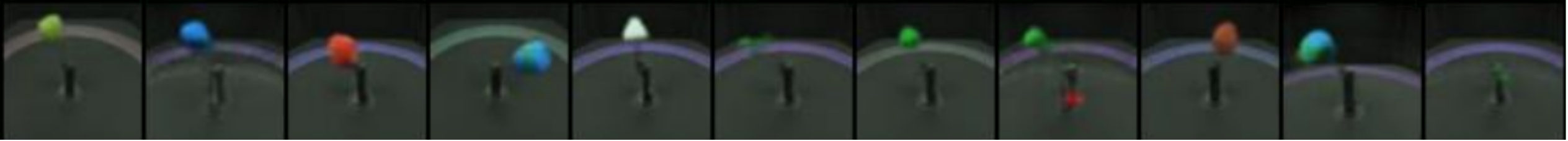}
      \put(0.6,1.9){\rotatebox{90}{\tiny{\textcolor{white}{F-VAE}}}}
    \end{overpic}%
    \vspace{-0.07\baselineskip}
    \begin{overpic}[width=\linewidth]{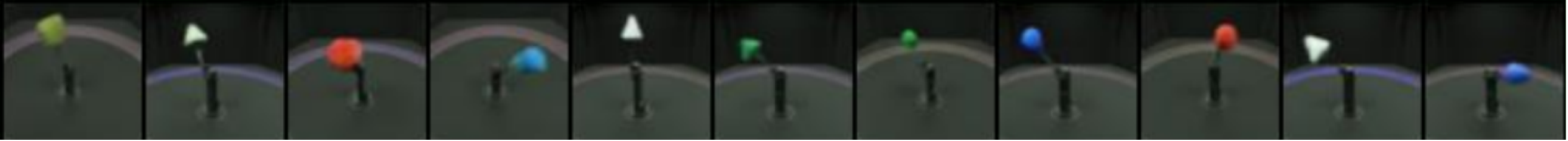}
      \put(0.6,2.1){\rotatebox{90}{\tiny{\textcolor{white}{E-AE}}}}
    \end{overpic}%
    \caption{Reconstruction of real data when being trained on realistic data.}
    \label{fig:mpi3d-realistic}
\end{subfigure}%
\vspace{0.5\baselineskip}
\begin{subfigure}{0.50\textwidth}
    \centering
    \begin{overpic}[width=\linewidth]{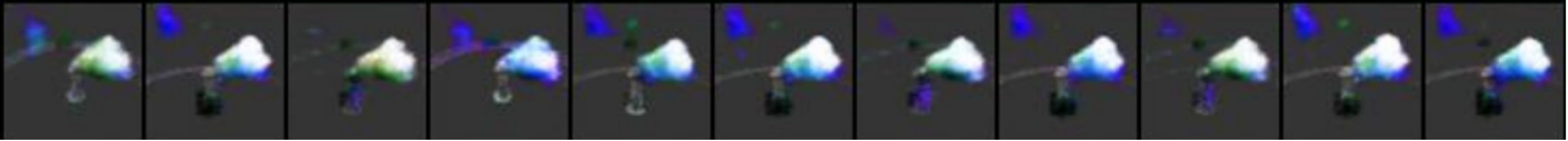}
      \put(0.5,2.9){\rotatebox{90}{\tiny{\textcolor{white}{AE}}}}
    \end{overpic}%
    \vspace{-0.07\baselineskip}
    \begin{overpic}[width=\linewidth]{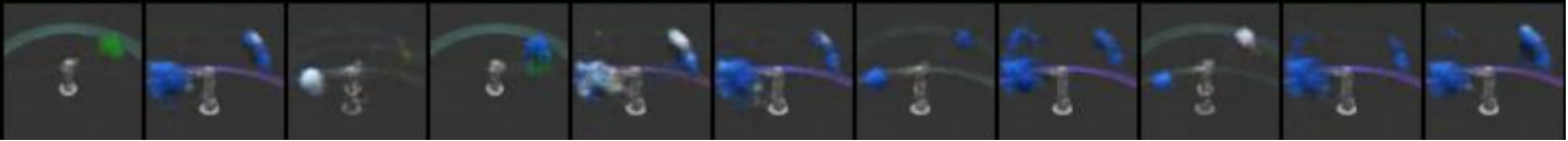}
      \put(0.5,1.9){\rotatebox{90}{\tiny{\textcolor{white}{$\beta$-AE}}}}
    \end{overpic}%
    \vspace{-0.07\baselineskip}
    \begin{overpic}[width=\linewidth]{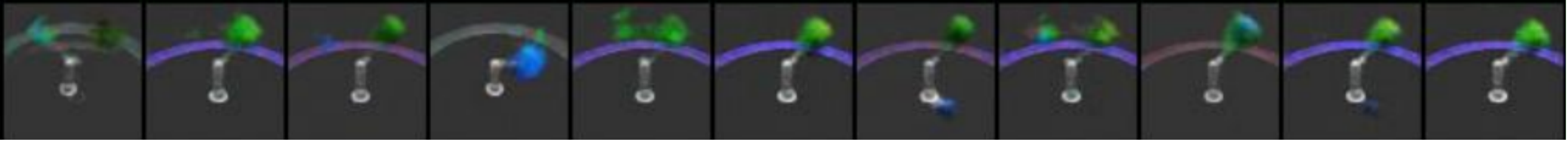}
      \put(0.5,1.9){\rotatebox{90}{\tiny{\textcolor{white}{F-VAE}}}}
    \end{overpic}%
    \vspace{-0.07\baselineskip}
    \begin{overpic}[width=\linewidth]{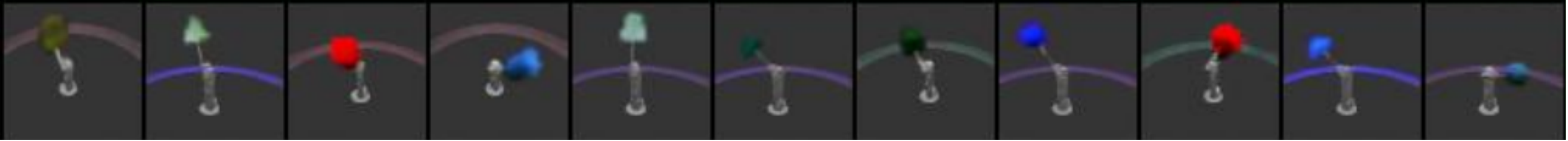}
      \put(0.5,2.1){\rotatebox{90}{\tiny{\textcolor{white}{E-AE}}}}
    \end{overpic}
    \caption{Reconstruction of real data when being trained on toy data.}
    \label{fig:mpi3d-toy}
\end{subfigure}
\caption{Reconstruction of unseen real data for different autoencoders: Autoencoder (AE), $\beta $ Variational Autoencoder ($\beta$-VAE), FactorVAE (F-VAE), Extractor Autoencoder (E-AE). Our methods preserves the semantics best.}
\label{fig:result-mpi3d}
\end{figure}

\begin{figure}
\centering
\begin{subfigure}{0.11\textwidth}
\centering
    \includegraphics[height=2.2cm]{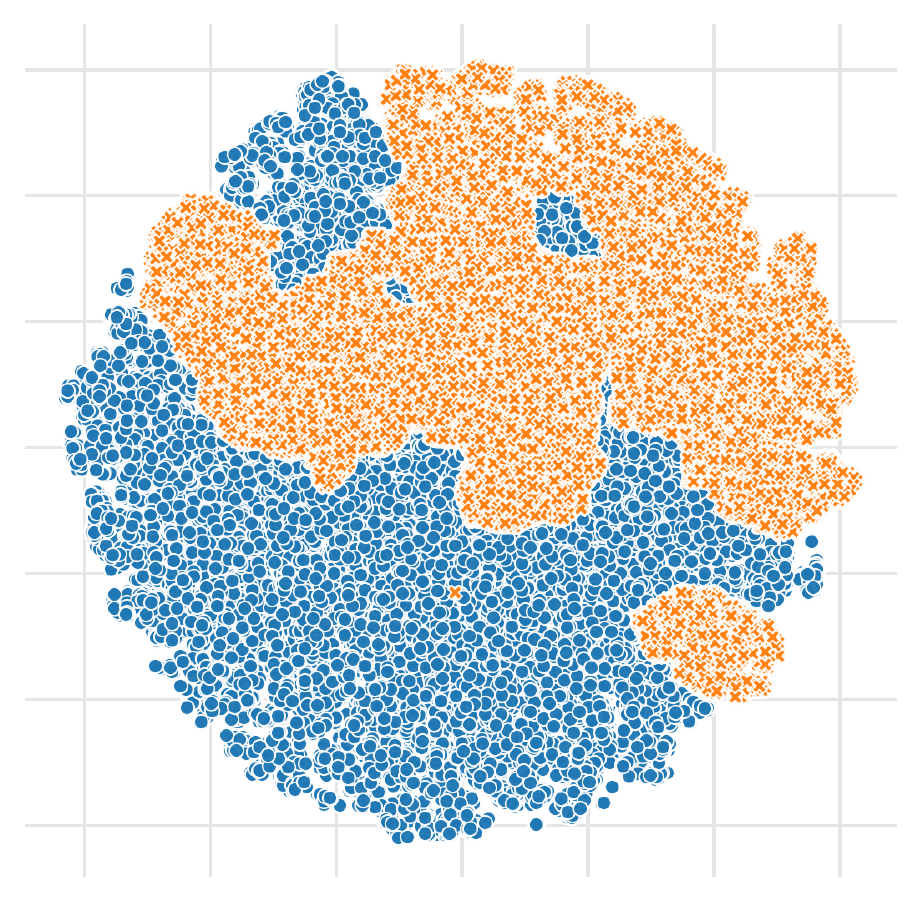}
    \caption{AE}
    \label{fig:latent-ae}
\end{subfigure}
\hfill
\begin{subfigure}{0.11\textwidth}
\centering
    \includegraphics[height=2.2cm]{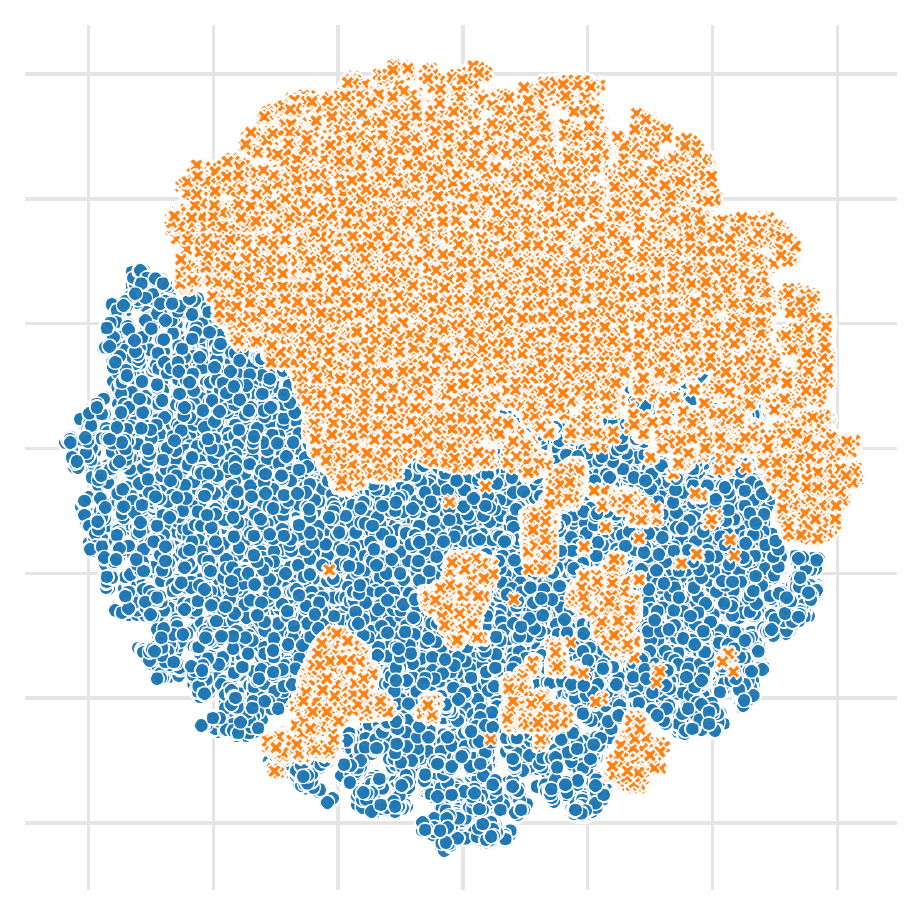}
    \caption{$\beta$-VAE}
    \label{fig:latent-beta-vae}
\end{subfigure}
\hfill
\begin{subfigure}{0.11\textwidth}
  \centering
      \includegraphics[height=2.2cm]{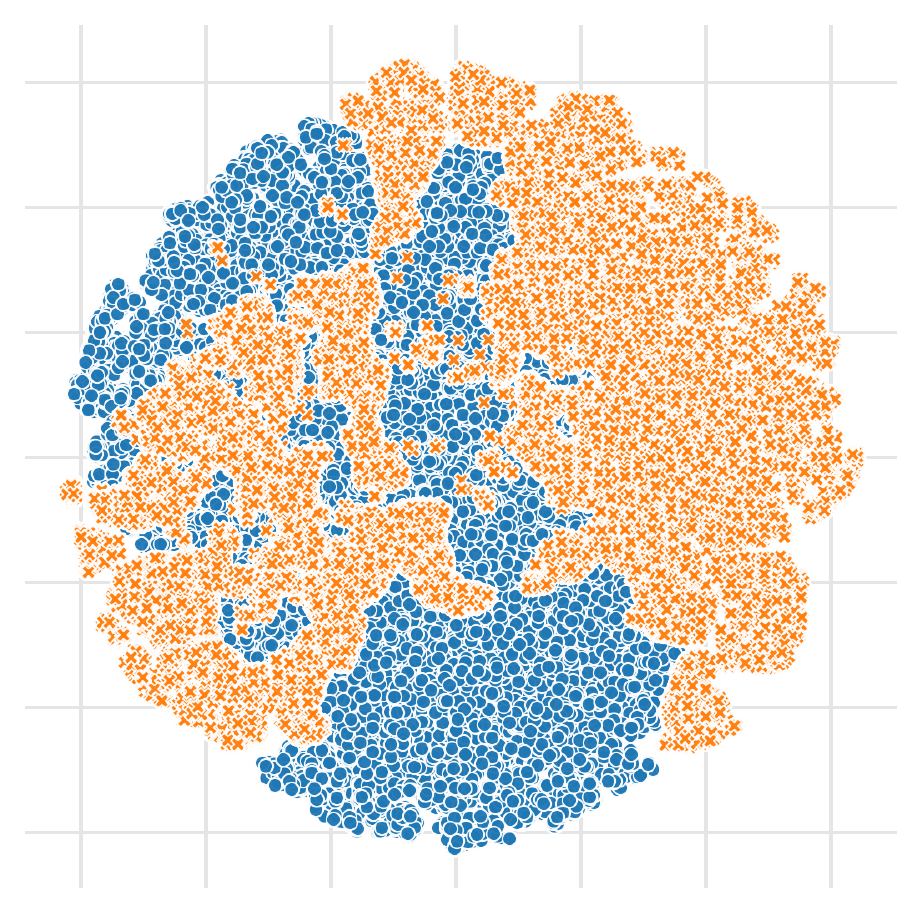}
      \caption{FactorVAE}
      \label{fig:latent-factor-vae}
  \end{subfigure}
  \hfill
\begin{subfigure}{0.11\textwidth}
  \centering
      \includegraphics[height=2.2cm]{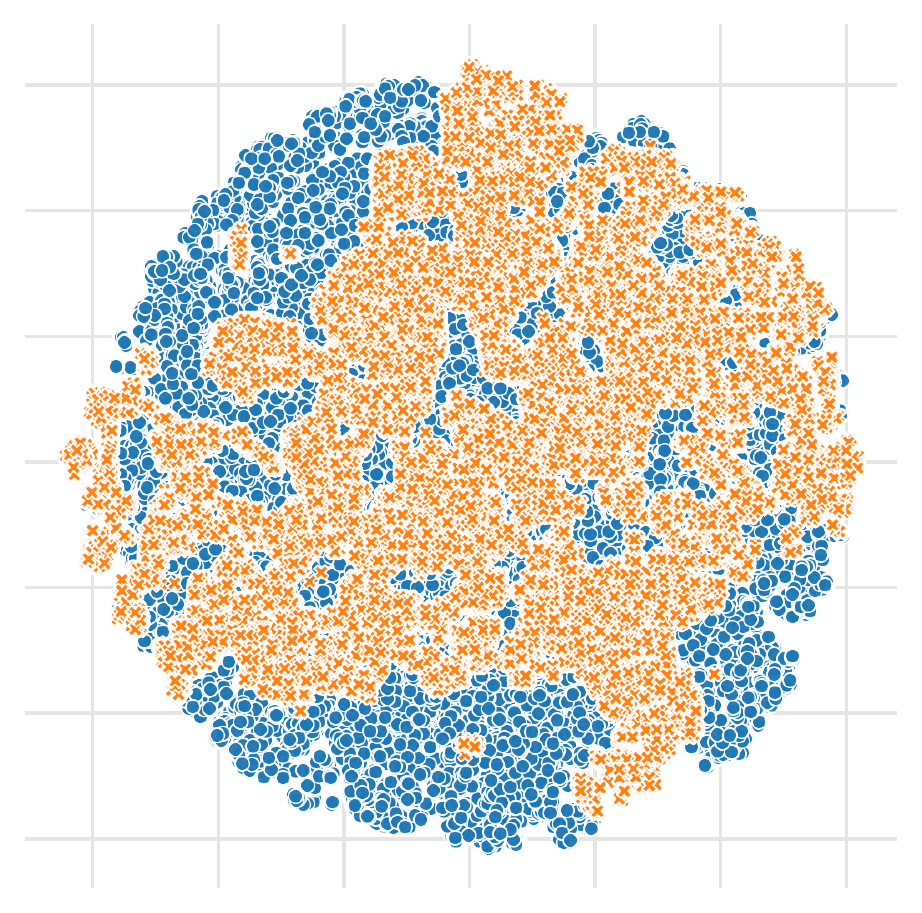}
      \caption{E-AE}
      \label{fig:latent-e-ae}
  \end{subfigure}
\caption{t-SNE projection of the 10 dimensional latent space representation of the realistic training (blue circle) together with the real (orange cross) images. Autoencoder (AE), $\beta $ Variational Autoencoder ($\beta$-VAE), FactorVAE and Extractor Autoencoder (E-AE). The extractor approach is the only method clustering both synthetic and real images together.}
\label{fig:latent}
\end{figure}

The results from the fourth row of Fig. \ref{fig:mpi3d-realistic} and Fig. \ref{fig:mpi3d-toy}, respectively, show that the proposed modifications enable the model to generalize to real images when trained on synthetic ones. Much more of the semantics are preserved even when the model was only trained on toy images. Our method produces semantically more correct and less noisy reconstructions compared to the VAE and FactorVAE baseline results. Additional qualitative improvements are highlighted by visualizing the latent space: both the $10$-dimensional training (synthetic) and test (real) data latent spaces are projected together into a $2$-dimensional representation using t-SNE. In Fig. \ref{fig:latent} we can observe that VAE and FactorVAE improve the representation of real and synthetic images in the same region in the latent space, however, only partially, indicating a different representation for real and synthetic images. When using E-AE, real and synthetic images are represented more similarly in the latent space and the clusters are completely overlapping. Even when trained on the toy dataset, the latent space representation for synthetic and real images produced by E-AE overlaps partially as visualized in the appendix Fig. S2. Finally, we report in Table \ref{table:mpi3d} a quantitative evaluation between the reconstructions of the real images against their synthetic training counterparts across all dataset images for different norms. We compute the same metrics between the real input images and their reconstruction to measure whether the semantics are being preserved : in all cases E-AE performs best. Additional results can be found in the appendix in Table S5 and reconstructions of synthetic input images in Fig S3. The latter shows that all models perform similarly well on the training data, hence the training was successful, but our proposed design choices generalize best to the real images.

\begin{table}
\caption{We report the SSIM and LPIPS \cite{zhang2018perceptual} norm between the reconstructions of the real images (unknown) and the corresponding synthetic (Synth.) training images (realistic (R) or toy (T)) or input images (Real). We report the mean of the norms across the dataset: for SSIM larger $\uparrow$ and for LIPIPS smaller $\downarrow$ is better. E-AE performs best. }
\label{table:mpi3d}
\centering
\begin{tabular}{rrrcccc}
  \toprule
  & & & \multicolumn{2}{c}{SSIM $\uparrow$}  &  \multicolumn{2}{c}{LPIPS $\downarrow$}  \\
   & Model & Variant & \multicolumn{1}{c}{Synth.} & \multicolumn{1}{c}{Real} & \multicolumn{1}{c}{Synth.} & \multicolumn{1}{c}{Real} \\
  \midrule
  \small{T} &  AE & SSIM  & 0.56 & 0.42 & 0.35 & 0.40 \\
  \small{T} &  VAE & BCE  & 0.50 & 0.33 & 0.34 & 0.42 \\
  \small{T} &  $\beta$-VAE & BCE, $\beta=4$ & 0.53 & 0.38 & 0.31 & 0.44 \\
  \small{T} &  $\beta$-VAE & BCE, $\beta=8$ & 0.71 & 0.48 & 0.26 & 0.37 \\
  \small{T} &  FactorVAE & BCE, $\gamma=10$ & 0.66 & 0.45 & 0.26 & 0.39 \\
  \small{T} &  FactorVAE & BCE, $\gamma=50$ & 0.71 & 0.51 & 0.22 & 0.35 \\
  \small{T} &  E-AE (ours) & SSIM & \cellcolor{good}0.90 & \cellcolor{good}0.58 & \cellcolor{good}0.10 & \cellcolor{good}0.2 \\
  \midrule
  \small{R} &  AE & SSIM & 0.83 & 0.62 & 0.20 & 0.24 \\
  \small{R} &  VAE & BCE & 0.74 & 0.61 & 0.20 & 0.23 \\
  \small{R} &  $\beta$-VAE & BCE, $\beta=4$ & 0.81 & 0.64 & 0.18 & 0.20 \\
  \small{R} &  $\beta$-VAE & BCE, $\beta=8$ & 0.79 & 0.64 & 0.19 & 0.21 \\
  \small{R} &  FactorVAE & BCE, $\gamma=10$ & 0.88 & 0.68 & 0.15 & 0.19 \\
  \small{R} &  FactorVAE & BCE, $\gamma=50$ & 0.78 & 0.64 & 0.16 & 0.18 \\
  \small{R} &  E-AE (ours) & SSIM & \cellcolor{good}0.92 & \cellcolor{good}0.70 & \cellcolor{good}0.08 & \cellcolor{good}0.14 \\
  \bottomrule
\end{tabular}
\end{table}
   
\subsection{It works for visually simple images - More is needed on more complex data}
\label{sec:complex-data}
Since the method introduced in the previous section achieved good results, even when being trained on toy images, we wanted to apply it to images of higher visual complexity, e.g. a vehicle interior. We trained the same model architecture, but with a 64-dimensional latent space, on images from the Tesla vehicle from SVIRO and the Kodiaq vehicle from SVIRO-Illumination, respectively, and evaluated the model on the real TICaM images. Examples of the resulting model's reconstructions are plotted in Fig. \ref{fig:result-reconstruction} (b) and in the appendix Fig. S4. In both cases only blurry human models are reconstructed, which is similar to the mode collapse in the first row of Fig. \ref{fig:mpi3d-toy}. We concluded that more robust features are needed.
\begin{figure}
  \begin{overpic}[width=0.99\linewidth]{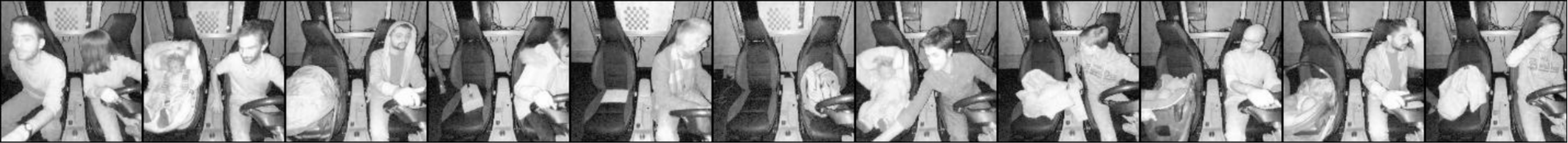}
      \put(0.5,3.7){\small{\textcolor{white}{(a)}}}
  \end{overpic}%
  \vspace{-0.07\baselineskip}
  \begin{overpic}[width=0.99\linewidth]{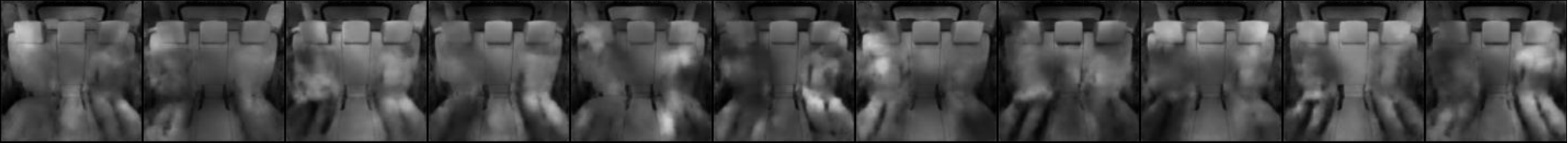}
      \put(0.5,3.7){\small{\textcolor{white}{(b)}}}
    \put(0.1,0.125){\color{Maroon}\framebox(8.9,8.9){}}
    \put(9.2 ,0.125){\color{ForestGreen}\framebox(8.9,8.9){}}
    \put(18.3,0.125){\color{Maroon}\framebox(8.9,8.9){}}
    \put(27.4 ,0.125){\color{Maroon}\framebox(8.85,8.9){}}
    \put(36.45,0.125){\color{Maroon}\framebox(8.9,8.9){}}
    \put(45.55 ,0.125){\color{Maroon}\framebox(8.9,8.9){}}
    \put(54.65,0.125){\color{Maroon}\framebox(8.9,8.9){}}
    \put(63.75 ,0.125){\color{Maroon}\framebox(8.85,8.9){}}
    \put(72.8,0.125){\color{Maroon}\framebox(8.9,8.9){}}
    \put(81.9 ,0.125){\color{Maroon}\framebox(8.9,8.9){}}
    \put(91.0,0.125){\color{Maroon}\framebox(8.85,8.9){}}
  \end{overpic}%
  \vspace{-0.07\baselineskip}
  \begin{overpic}[width=0.99\linewidth]{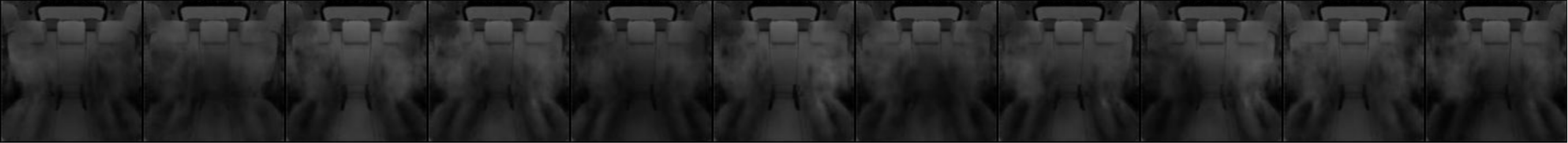}
      \put(0.5,3.7){\small{\textcolor{white}{(c)}}}
    \put(0.1,0.125){\color{Maroon}\framebox(8.9,8.9){}}
    \put(9.2 ,0.125){\color{Maroon}\framebox(8.9,8.9){}}
    \put(18.3,0.125){\color{Maroon}\framebox(8.9,8.9){}}
    \put(27.4 ,0.125){\color{Maroon}\framebox(8.85,8.9){}}
    \put(36.45,0.125){\color{Maroon}\framebox(8.9,8.9){}}
    \put(45.55 ,0.125){\color{Maroon}\framebox(8.9,8.9){}}
    \put(54.65,0.125){\color{Maroon}\framebox(8.9,8.9){}}
    \put(63.75 ,0.125){\color{Maroon}\framebox(8.85,8.9){}}
    \put(72.8,0.125){\color{Maroon}\framebox(8.9,8.9){}}
    \put(81.9 ,0.125){\color{Maroon}\framebox(8.9,8.9){}}
    \put(91.0,0.125){\color{Maroon}\framebox(8.85,8.9){}}
  \end{overpic}%
  \vspace{-0.07\baselineskip}
  \begin{overpic}[width=0.99\linewidth]{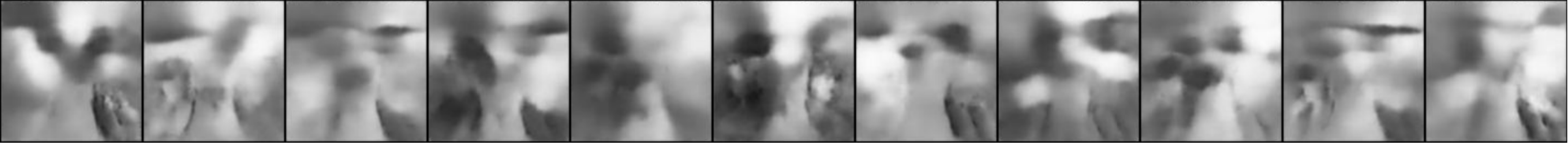}
      \put(0.5,3.7){\small{\textcolor{white}{(d)}}}
    \put(0.1,0.125){\color{Maroon}\framebox(8.9,8.9){}}
    \put(9.2 ,0.125){\color{Maroon}\framebox(8.9,8.9){}}
    \put(18.3,0.125){\color{Maroon}\framebox(8.9,8.9){}}
    \put(27.4 ,0.125){\color{Maroon}\framebox(8.85,8.9){}}
    \put(36.45,0.125){\color{Maroon}\framebox(8.9,8.9){}}
    \put(45.55 ,0.125){\color{Maroon}\framebox(8.9,8.9){}}
    \put(54.65,0.125){\color{Maroon}\framebox(8.9,8.9){}}
    \put(63.75 ,0.125){\color{Maroon}\framebox(8.85,8.9){}}
    \put(72.8,0.125){\color{Maroon}\framebox(8.9,8.9){}}
    \put(81.9 ,0.125){\color{Maroon}\framebox(8.9,8.9){}}
    \put(91.0,0.125){\color{Maroon}\framebox(8.85,8.9){}}
  \end{overpic}%
  \vspace{-0.07\baselineskip}
  \begin{overpic}[width=0.99\linewidth]{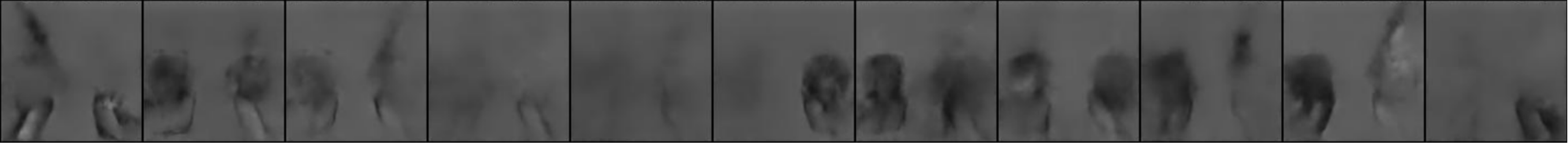}
      \put(0.5,3.7){\small{\textcolor{white}{(e)}}}
    \put(0.1,0.125){\color{Maroon}\framebox(8.9,8.9){}}
    \put(9.2 ,0.125){\color{ForestGreen}\framebox(8.9,8.9){}}
    \put(18.3,0.125){\color{Maroon}\framebox(8.9,8.9){}}
    \put(27.4 ,0.125){\color{Maroon}\framebox(8.85,8.9){}}
    \put(36.45,0.125){\color{Maroon}\framebox(8.9,8.9){}}
    \put(45.55 ,0.125){\color{Maroon}\framebox(8.9,8.9){}}
    \put(54.65,0.125){\color{ForestGreen}\framebox(8.9,8.9){}}
    \put(63.75 ,0.125){\color{Maroon}\framebox(8.85,8.9){}}
    \put(72.8,0.125){\color{Maroon}\framebox(8.9,8.9){}}
    \put(81.9 ,0.125){\color{ForestGreen}\framebox(8.9,8.9){}}
    \put(91.0,0.125){\color{Maroon}\framebox(8.85,8.9){}}
  \end{overpic}%
  \vspace{-0.07\baselineskip}
  \begin{overpic}[width=0.99\linewidth]{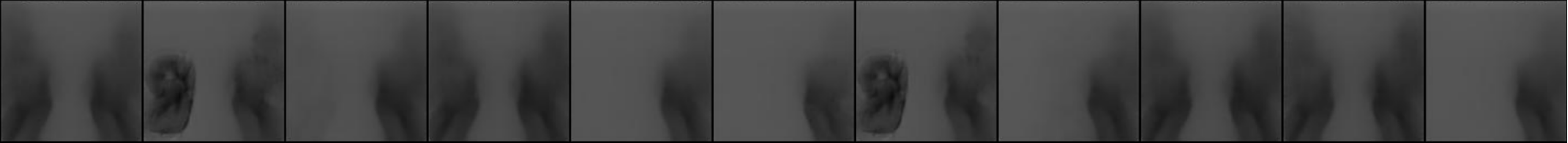}
    \put(0.5,3.7){\small{\textcolor{white}{(f)}}}
    \put(0.1,0.125){\color{ForestGreen}\framebox(8.9,8.9){}}
    \put(9.2 ,0.125){\color{ForestGreen}\framebox(8.9,8.9){}}
    \put(18.3,0.125){\color{Maroon}\framebox(8.9,8.9){}}
    \put(27.4 ,0.125){\color{Maroon}\framebox(8.85,8.9){}}
    \put(36.45,0.125){\color{ForestGreen}\framebox(8.9,8.9){}}
    \put(45.55 ,0.125){\color{Maroon}\framebox(8.9,8.9){}}
    \put(54.65,0.125){\color{ForestGreen}\framebox(8.9,8.9){}}
    \put(63.75 ,0.125){\color{ForestGreen}\framebox(8.85,8.9){}}
    \put(72.8,0.125){\color{Maroon}\framebox(8.9,8.9){}}
    \put(81.9 ,0.125){\color{Maroon}\framebox(8.9,8.9){}}
    \put(91.0,0.125){\color{ForestGreen}\framebox(8.85,8.9){}}
  \end{overpic}%
  \vspace{-0.07\baselineskip}
  \begin{overpic}[width=0.99\linewidth]{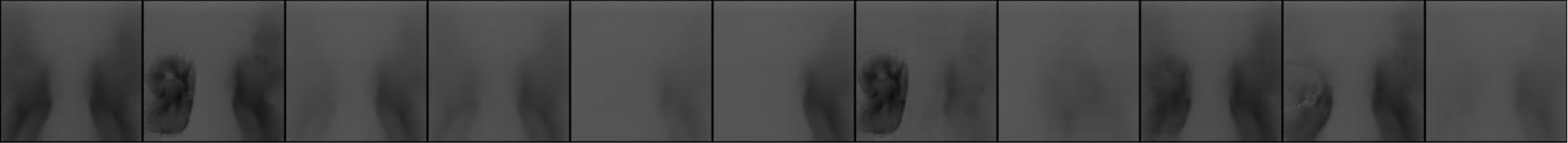}
      \put(0.5,3.7){\small{\textcolor{white}{(g)}}}
    \put(0.1,0.125){\color{ForestGreen}\framebox(8.9,8.9){}}
    \put(9.2 ,0.125){\color{ForestGreen}\framebox(8.9,8.9){}}
    \put(18.3,0.125){\color{Maroon}\framebox(8.9,8.9){}}
    \put(27.4 ,0.125){\color{Maroon}\framebox(8.85,8.9){}}
    \put(36.45,0.125){\color{ForestGreen}\framebox(8.9,8.9){}}
    \put(45.55 ,0.125){\color{Maroon}\framebox(8.9,8.9){}}
    \put(54.65,0.125){\color{ForestGreen}\framebox(8.9,8.9){}}
    \put(63.75 ,0.125){\color{Maroon}\framebox(8.85,8.9){}}
    \put(72.8,0.125){\color{Maroon}\framebox(8.9,8.9){}}
    \put(81.9 ,0.125){\color{ForestGreen}\framebox(8.9,8.9){}}
    \put(91.0,0.125){\color{ForestGreen}\framebox(8.85,8.9){}}
  \end{overpic}%
  \caption{Reconstructions of unseen real data (a) from TICaM: (b) E-AE and (c) I-E-AE trained on Kodiaq SVIRO-Illumination, (d) E-AE, (e) I-E-AE, (f) II-E-AE and (g) II-E-TAE trained on our new dataset. A red (wrong) or green (correct) box highlights whether the classes are preserved.}
  \label{fig:result-reconstruction}
\end{figure}
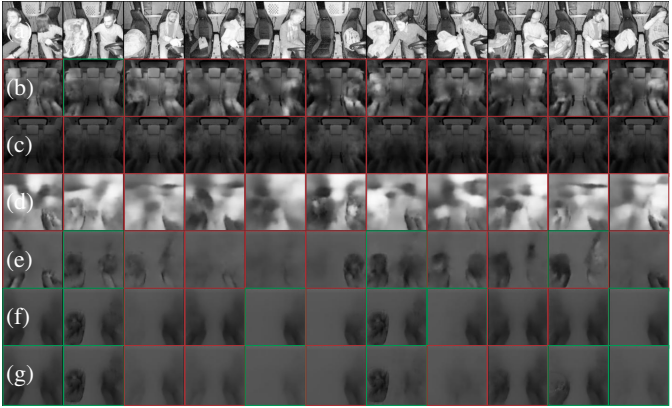

\subsection{PIRL helps generalization}
\label{sec:impossible}
As defined in Eq. \eqref{eq:recon-loss-I}, a partially impossible reconstruction loss (PIRL) for autoencoders has proven to work well for image normalization \cite{DiasDaDa2021Illumination}. We hypothesized that the same approach could lead to a better generalization to real vehicle interiors. We applied this strategy to variations of the same scene under different illumination conditions, but realized that the learned invariances are not suitable for the transfer between synthetic and real. An example is provided in Fig. \ref{fig:result-reconstruction} (c) where we trained on the Kodiaq images from SVIRO-Illumination. 

We concluded that, for learning more general features by applying the PIRL, we needed input-target pairs where both images are of the same scene, but differ in the properties we want to become invariant to: the dominant background. To this end we created 5919 synthetic scenes where we placed humans, child and infant seats as if they would be sitting in a vehicle interior, but instead of a vehicle, the background was replaced by selecting randomly from a pool of available HDRI images. Each scene was rendered using 10 different backgrounds. Examples from the dataset are shown in Fig. S1 in the appendix. During training, we randomly select two images per scene and use one as input and the other as target, i.e. as defined in Eq. \eqref{eq:recon-loss-I}. When applied to real images, see Fig. \ref{fig:result-reconstruction} (e), the model better preserves the semantics of the real images: the model starts to reconstruct child seats and not people only, anymore. We also trained a model without the PIRL to show that the success is not due to the design choice of the dataset: in Fig. \ref{fig:result-reconstruction} (d) the model performs worse.

Finally, we extended this idea further with our novel PIRL loss variation: instead of taking the same scene with a different background as target image, we randomly selected a different scene of the same class, e.g. if a person is sitting at the left seat position, we take another image with a person on the left seat, potentially a different person with a different pose. This approach is formulated in Eq. \eqref{eq:recon-loss-II}. While this leads to a blurrier reconstruction, which is expected because the autoencoder needs to learn an average class representation, the classes are preserved more robustly and the reconstructions look better than before, see Fig. \ref{fig:result-reconstruction} (f). This additional randomization improves classification accuracy as discussed in Sections \ref{sec:classification} and \ref{sec:limitation}. A visualization of the different input-target pairs can be found in Fig. \ref{fig:input} and the dataset \href{https://sviro.kl.dfki.de/sviro-nocar-download/}{can been downloaded (link)}.
   
\subsection{Structure in the latent space helps generalization}
\label{sec:structure}
The final improvement is based on the assumption that structure in the latent space should help the model performance. Class labels are included by formulating a triplet loss regularization to the latent space representation as defined by Eq. \eqref{eq:triplet-loss}: images of the same class should be mapped closely together and images of different classes pushed away. The triplet loss induces a more meaningful $L^2$-norm in the latent space \cite{DiasDaDa2021Illumination} such that a k-nearest neighbour (KNN) classifier can be used in the next section. As the results of Fig. \ref{fig:result-reconstruction} (g) and in the appendix show, these final improvements, together with the previous changes, yield the semantically most correct reconstructions. In the appendix we show that due to the triplet loss the nearest neighbour of (g) makes sense and yields a clearer reconstruction. The triplet loss without the PIRL is not sufficient and in Section \ref{sec:limitation} we show that the II-PIRL loss is the driving force for the improved performance.

\subsection{KNN with triplet loss out-performs classification models}
\label{sec:classification}
\begin{figure}
  \centering
  \begin{subfigure}{0.46\textwidth}
  \centering
      \includegraphics[width=\textwidth]{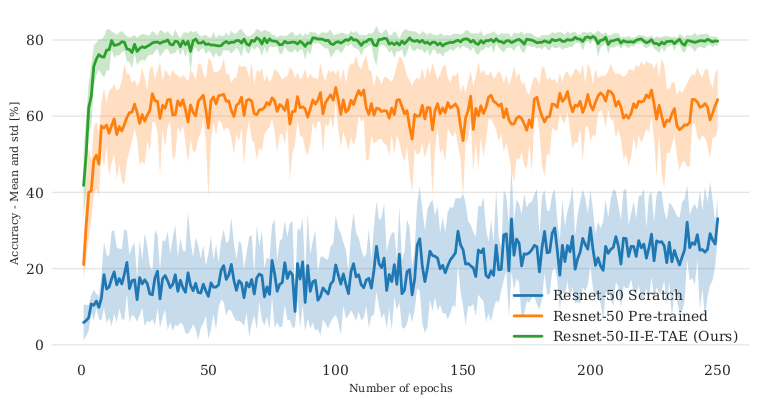}
      \caption{Resnet-50}
  \end{subfigure}%
  \hfill
  \begin{subfigure}{0.46\textwidth}
  \centering
      \includegraphics[width=\textwidth]{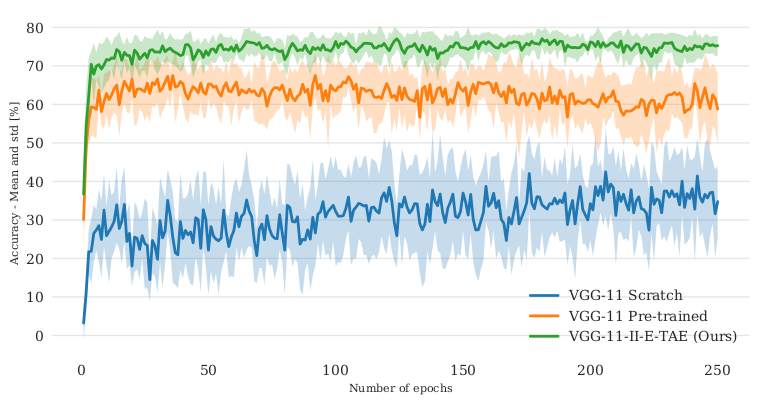}
      \caption{VGG-11}
  \end{subfigure}
  \caption{Training performance distribution for each epoch over 250 epochs. II-E-TAE is compared against training the corresponding extractor from scratch or fine-tuning the layers after the features used by the extractor.}
  \label{fig:accuracy}
\end{figure}

We investigated whether the qualitative improvements also transfer to a quantitative improvement. We took the most basic approach: we combined the E-TAE with a k-nearest neighbour classifier in the latent space and used our new dataset for training. We retrieve the latent space vectors for all flipped training images as well and used only a single image per scene (i.e. not all 10 variations). We choose $k=\sqrt{N}=115$, where $N$ is the size of the training data together with its flipped version \cite{jirina2011classifiers}. The model should classify occupancy (empty, infant, child or adult) for each seat position and we used the same hyperparameters for all methods and variations thereof. We froze the same layers of the pre-trained models for fine-tuning the later layers in case of classification models or to train our autoencoder using it as an extractor. We evaluated the model performance after each epoch on the real TICaM images (normal and flipped images of the training and test splits) for both the autoencoder and the corresponding classification model. This provides a measure on the best possible result for each method, but is of course not a valid approach for model selection. We report in Fig. \ref{fig:accuracy} the training results for seeds 1 to 10 and summarize the training performance by plotting the mean and standard deviation per epoch per method. Our approach converges more robustly and consistently to a better mean accuracy. For each experiment, we retrieve the best accuracy across all epochs and compute the mean, standard deviation and maximum of these values across all runs: these statistics are reported in Table \ref{table:accuracy}. See the appendix for training from scratch and Densenet-121 results. The model weights corresponding to the epochs selected by the previous heuristics were applied on the SVIRO dataset to verify whether the learned representations are universally applicable to other vehicle interiors. For SVIRO, we used the training images and excluded all images containing empty child seats or empty infant seats, treated everyday objects as background. The results show that our E-AE significantly outperforms the classification models across three different pre-trained models and across all datasets. A consistent improvement for the different modifications is achieved: I-E-TAE outperforms E-TAE and II-E-TAE outperforms I-E-TAE. 
\begin{table}
  \caption{For each experiment, the best accuracy on real TICaM images across all epochs is taken and the mean, standard deviation and maximum of those values across all 10 runs is reported. The model weights achieving maximum performance per run on TICaM are evaluated on SVIRO.}
  \label{table:accuracy}
  \centering
  \begin{tabular}{lrcccc}
    \toprule
    &  & \multicolumn{2}{c}{TICaM} & \multicolumn{2}{c}{SVIRO} \\
    \midrule
    Model & \multicolumn{1}{c}{Variant} & \multicolumn{1}{c}{Mean} & \multicolumn{1}{c}{Max} & \multicolumn{1}{c}{Mean} & \multicolumn{1}{c}{Max} \\
    \midrule
    \small{VGG} & \small{Pretrained} & $75.5 \pm 1.5$ & $78.0$ & $78.7 \pm 2.9$ & $84.0$ \\
    \small{Resnet} & \small{Pretrained}  &  $78.1 \pm 1.7$ & $80.4$ &  $83.5 \pm 2.7$ & $88.1$  \\
    \midrule
    \small{VGG} &  \small{E-TAE} & $76.7 \pm 2.3$  & $81.5$ &  $78.6 \pm 2.6$ & $82.3$\\
    \small{Resnet} &  \small{E-TAE} & \cellcolor{good}$83.8 \pm 1.3$  & \cellcolor{good}86.0 &  $85.8 \pm 2.4$ & $89.1$ \\
    \midrule
    \small{VGG} &  \small{I-E-TAE} & $79.7 \pm 2.1$  & \cellcolor{good}82.2 &  \cellcolor{good}$80.9 \pm 4.0$ & \cellcolor{good}85.6 \\
    \small{Resnet} &  \small{I-E-TAE} & $83.5 \pm 1.3$  & $85.6$ &  $89.2 \pm 1.0$ & $90.3$ \\
    \midrule
  \small{VGG} &  \small{II-E-TAE} & \cellcolor{good}$81.0 \pm 0.6$  & $82.0$ &  $79.1 \pm 3.9$ & $84.8$ \\
    \small{Resnet} & \small{II-E-TAE} & \cellcolor{good}$83.7 \pm 0.5$  & $84.5$ &  \cellcolor{good}$93.0 \pm 0.8$ & \cellcolor{good}94.1\\
    \bottomrule
  \end{tabular}
\end{table}

\begin{table}
  \caption{For each of the 10 runs per method after 250 epochs using the VGG-11 extractor we trained different classifiers in the latent space: k-nearest neighbour (KNN), random forest (RForest) and support vector machine with a linear kernel (SVM). Most of the contribution to the synthetic to real generalization is due to the novel II variation of the PIRL.}
  \label{table:classifier-comparison}
  \centering
  \begin{tabular*}{0.48\textwidth}{@{\extracolsep{\fill}}rccc}
    \toprule
    Variant & KNN & RForest & SVM \\
    \midrule
    \small{E-AE} & $17.1 \pm 6.7$  & $24.2 \pm 4.1$ & $40.6 \pm 8.5$\\
    \small{I-E-AE} & $18.2 \pm 7.3$ & $42.4 \pm 6.5$ & $50.1 \pm 3.7$ \\
    \small{II-E-AE} & \cellcolor{good}$73.2 \pm 3.9$ & \cellcolor{good}$68.8 \pm 5.7$ & $66.9 \pm 6.7$  \\
    \small{E-TAE} & $69.2 \pm 3.4$ & $66.4 \pm 4.0$ & \cellcolor{good}$68.7 \pm 2.2$  \\
    \bottomrule
  \end{tabular*}
\end{table}

\section{Discussion and Limitations}
\label{sec:limitation}
We want to highlight that most of the contribution to the success of our introduced model variations stems from the novel II variation of the PIRL loss. To this end we trained several types of classifiers in the latent space of different autoencoder model variations and report the results in Table \ref{table:classifier-comparison}. The II variation of the PIRL loss largely improves the classification accuracy compared to the I variation. Moreover, the performance is better compared to the triplet loss variation which uses the label information explicitly as a latent space constraints, compared to the implicit use by the II-PIRL.

The II variation of the PIRL loss implicitly assumes that the classes are uni-modal, i.e. objects of the same class should be mapped onto a similar point in the latent space. This characteristic can either improve generalization or have a detrimental effect on the performance depending on the task to be solved. Under its current form there is no guarantee that, for example, facial landmarks or poses would be preserved. Nevertheless, we believe that extensions of our proposed loss, for example based on constraints (e.g. preservation of poses) could be an interesting direction for future work. It can be observed that our model is not perfect and sometimes struggles: e.g. for more complex human poses (e.g. people turning over). However, we believe that these problems are related to the training data: a more versatile synthetic dataset would probably improve the model performance on more challenging real images. 

Finally, we show that improvements reported in this work are not limited to the application in the vehicle interior. To this end, we trained models using the same design choices on MNIST \cite{lecun1998gradient} and evaluate the generalization onto real digits \cite{de2009character} in Fig. S6 and Table S7 in the appendix: similar improvements by the different design choices can be observed.

\section{Conclusion}
\label{sec:conclusion}
We introduced an autoencoder model which uses a pre-trained classification model as a feature extractor. Our results showed that the resulting model produces superior reconstructions for synthetic to real generalization. However, we highlighted that design choices made on simple datasets do not necessarily transfer to visually more complex tasks. We performed a step-by-step investigation of additional model changes and showcased the improvements of each change. Although a k-nearest neighbour classifier is used in the latent space, our proposed autoencoder model outperforms consistently and more robustly all classification model counterparts.

\section*{Acknowledgment}
The first author is supported by the Luxembourg National Research Fund (FNR) under grant number 13043281. The second author is supported by DECODE (grant number 01IW21001). This work was partially funded by the Luxembourg Ministry of the Economy (CVN 18/18/RED).

\newpage


\bibliographystyle{IEEEtran}
\bibliography{bibliography}

\end{document}